\documentclass[runningheads]{llncs}
\usepackage{graphicx}
\usepackage{multirow}
\usepackage{threeparttable}

\begin{document}
%
% \title{Contribution Title\thanks{Supported by organization x.}}
\title{CaFT: Clustering and Filter on Tokens of Transformer for Weakly Supervised Object Localization}
\titlerunning{CaFT: Clustering and Filter on Tokens of Transformer for WSOL}
% %
% \author{Ming Li\inst{1}\orcidID{0000-1111-2222-3333} \and
\author{Ming Li\inst{1}}
\institute{College of Control Science and Engineering, Zhejiang University, China
\email{22132077@zju.edu.cn}\\
}

\maketitle              % typeset the header of the contribution
\begin{abstract}
% The abstract should briefly summarize the contents of the paper in
% 15--250 words.
Weakly supervised object localization (WSOL) is a challenging task to localize the object by only category labels. However, there is contradiction between classification and localization because accurate classification network tends to pay attention to discriminative region of objects rather than the entirety. We propose this discrimination is caused by handcraft threshold choosing in CAM-based methods. Therefore, we propose Clustering and Filter of Tokens (CaFT) with Vision Transformer (ViT) backbone to solve this problem in another way. CaFT first sends the patch tokens of the image split to ViT and cluster the output tokens to generate initial mask of the object. Secondly, CaFT considers the initial mask as pseudo labels to train a shallow convolution head (Attention Filter, AtF) following backbone to directly extract the mask from tokens. Then, CaFT splits the image into parts, outputs masks respectively and merges them into one refined mask. Finally, a new AtF is trained on the refined masks and used to predict the box of object. Experiments verify that CaFT outperforms previous work and achieves 97.55\% and 69.86\% localization accuracy with ground-truth class on CUB-200 and ImageNet-1K respectively. CaFT provides a fresh way to think about the WSOL task.
\end{abstract}
\section{Introduction}
Current deep learning methods have achieved enormous development and wide-spread application. However, the fully supervised learning always needs a mass of accurate labeled data, which is costly to acquire. So weakly supervised learning become an important and challenging field to be explored. In recent years, many methods are proposed to solve WSOL tasks.

Weakly supervised object location (WSOL) aims to localize the object in an image without the label of its bounding box. The most important method of WSOL is the Class Activation Mapping (CAM)\cite{cam}. It utilizes the activation map from the last convolution layer to generate the bounding box of object. But the classification models prefer to pay attention to the most discriminative region of the object rather than the entirety, as well as many developed methods based on CAM \cite{has,acol,spg,adl}. Also, after the model generates the activation maps, the threshold of activation value should be chosen to get the final bounding box, which is a process with lots of handcraft design. Because there are different activation conditions in different images, it is difficult to set a threshold for all samples. 

Therefore, we propose a method utilizing k-means \cite{kmeans} clustering algorithm to realize automatically learning the threshold of each region in one image. By dividing the feature map from convolutional neural network (CNN) into eigenvectors with length of channels, we intend to cluster these vectors. However there is a problem to decide which cluster represents the region of object. Vision Transformer (ViT) \cite{vit} is a new framework with self-attention mechanism. It possesses the special class token, which is independent of other tokens in the output feature map. By putting the class token into clustering, it is easy to choose the cluster of class token as the foreground and generate the mask of the object.

In addition, following the previous work \cite{psol}, we also try to shift the weakly supervised task to a pseudo supervised task. The method of \cite{psol} generates pseudo bounding-box labels. Different from it, we consider the mask result of clustering as pseudo label and use a shallow convolutional module (called Attention Filter, AtF) following ViT’s output tokens to generate a mask more accurate and robust. Also, AtF is designed as a classification module with two classes (foreground and background) to directly output the region of object, which is no need to set a threshold in a regression method as well.

We summarize our contributions as follows:
\begin{enumerate}
  \item[-] We propose a novel combination of clustering and deep-learning in WSOL task, which helps to get rid of the difficult threshold-choosing in CAM-based method and generate a complete mask of object.
  \item[-] We propose a light-weight module, AtF, trained on pseudo mask labels, to extract information from feature map and generate an accurate mask of object.
  \item[-] We propose a high-quality transformer-based architecture to solve the WSOL task, which achieves 97.55\% and 69.86\% localization accuracy with ground-truth class on CUB-200 and ImageNet-1K respectively. We name the method as \textbf{C}lustering \textbf{a}nd \textbf{F}ilter on \textbf{T}okens (CaFT).
\end{enumerate}

\section{Related Work}
\subsubsection{Weakly Supervised Object Localization (WSOL)}  is a challenging task to learn object localizations by given only the category labels. It is attractive because image-level category labels are easier and more inexpensive to obtain than localization labels.

CAM \cite{cam} is the pioneer work in WSOL. It utilizes the global average pooling to replace the fully connected layer in CNN-based classification models, and utilizes the class activation maps to generate the region of object. However, there are problems to use class activation maps. Firstly, the class activation maps usually pay attention to the most discriminative parts of object. Secondly, after generating class activation maps, it needs to set a handcraft threshold to distinguish the foreground and background. Different choices of threshold often greatly influences the performance of bounding-box prediction. 

\subsubsection{Development of CAM-based Methods.}To solve the problems of CAM and improve the performance of localization, a lot of different methods are proposed. 

For example, HaS \cite{has} divides the image into several patches then inputs them into network during training, which aims to reduce the reliance on the most discriminative regions. CutMix \cite{cutmix} mixes patches of different images. ACoL \cite{acol} uses two branches of classifier to erasing and predict discriminative regions. ADL \cite{adl} tries to erase feature maps corresponding to discriminative regions during inference. These methods are based on erasing some information of images during training or inference. DA-Net \cite{danet} utilizes a discrepant activation method reducing the similarity of CAMs and gathers to a less discriminative activation maps. SPG \cite{spg} and I$^2$C \cite{i2c} restrict the pixel-level correlations in the network to reduce the dependence on discriminative regions. PSOL \cite{psol} finds that localization and classification should be divided into two separate tasks, and uses class-agnostic localization method to predict bounding boxes.

\subsubsection{Vision Transformer.} Transformer is initially an architecture to solve the sequential tasks. Recently, it is applied to domain of computer vision. ViT \cite{vit} uses the pure transformer directly to sequences of image patches for exploring spatial correlation. And it achieves a great performance on classification tasks. Deit \cite{deit} introduces several training strategies that allow ViT to be also effective when using the smaller ImageNet-1K dataset. The results of ViT on image classification are encouraging. Lately, there is work to deploy transformer architecture to WSOL tasks. TS-CAM \cite{tscam} firstly splits an image into a sequence of patch tokens for spatial embedding, then re-allocates category-related semantics for patch tokens, finally couples the patch tokens with the semantic-agnostic attention map to achieve semantic-aware localization.

All the methods mentioned above give a big boost on WSOL. But there is still weakness on the discriminative region and threshold choosing. Our CaFT replaces the handcraft threshold choosing to a self-learning process by clustering the points on the feature maps. Due to the label-agnostic clustering and the global vision of ViT, activated region is discretized and the threshold of foreground and background is no longer fixed. Therefore, it sets a more flexible boundary to segment the complete object from background rather than the region beneficial to classification. 

\begin{figure}
\includegraphics[width=\textwidth]{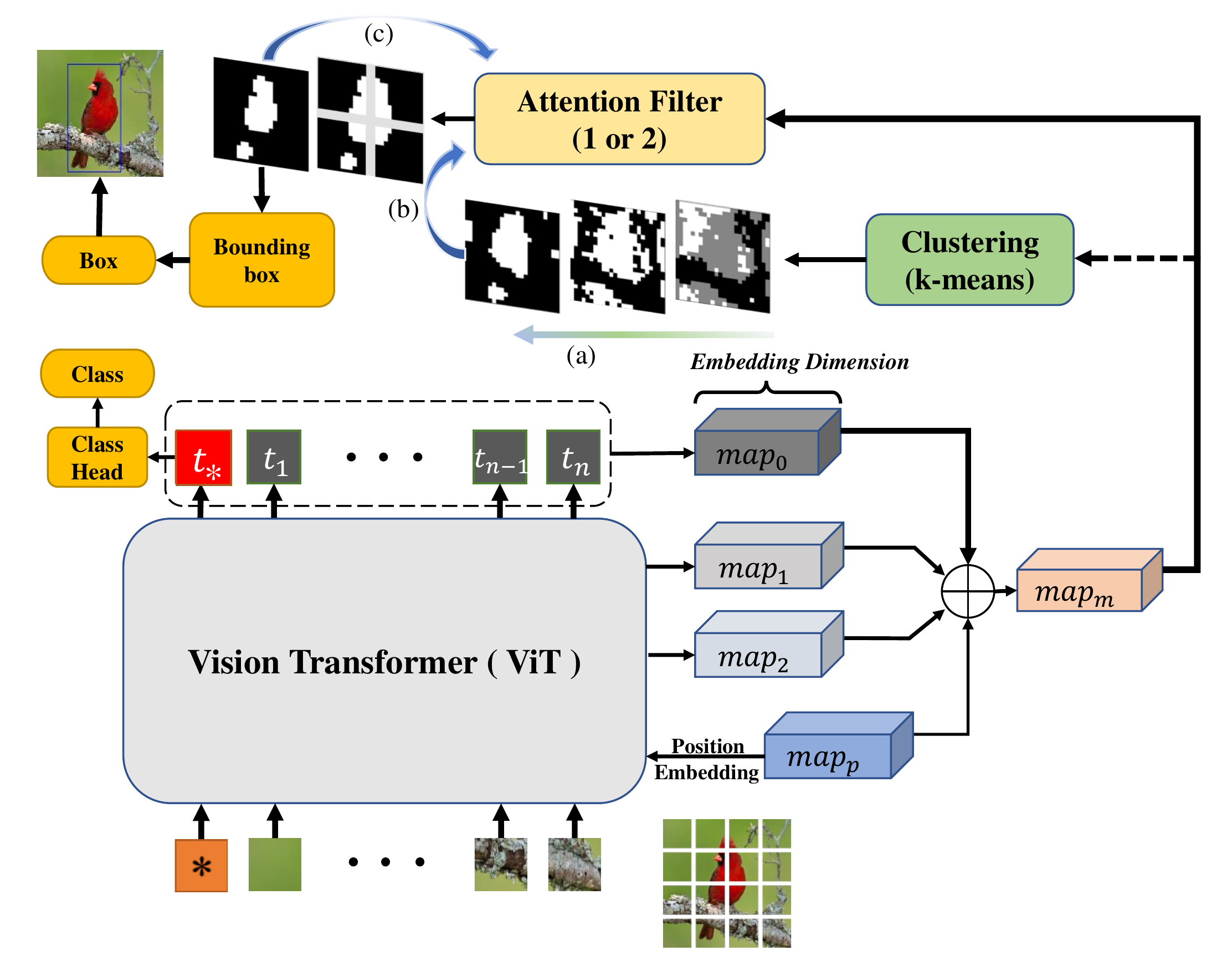}
\caption{CaFT structure. The ViT backbone converts the image to tokens and outputs the category. Then the last three layers of tokens and position embedding parameters are merged together to $map_m$. In training, (a) clustering tokens and postprocessing; (b) using the mask of (a) to train the AtF$_1$; (c) dividing the input image and merging the output masks to train the AtF$_2$. Then CaFT uses the output mask of AtF$_2$ to draw the bounding box. In inference, CaFT directly inputs the $map_m$ to AtF$_2$ to predict mask and the bounding box.} 
\label{structure}
\end{figure}

\begin{figure}
\includegraphics[width=\textwidth]{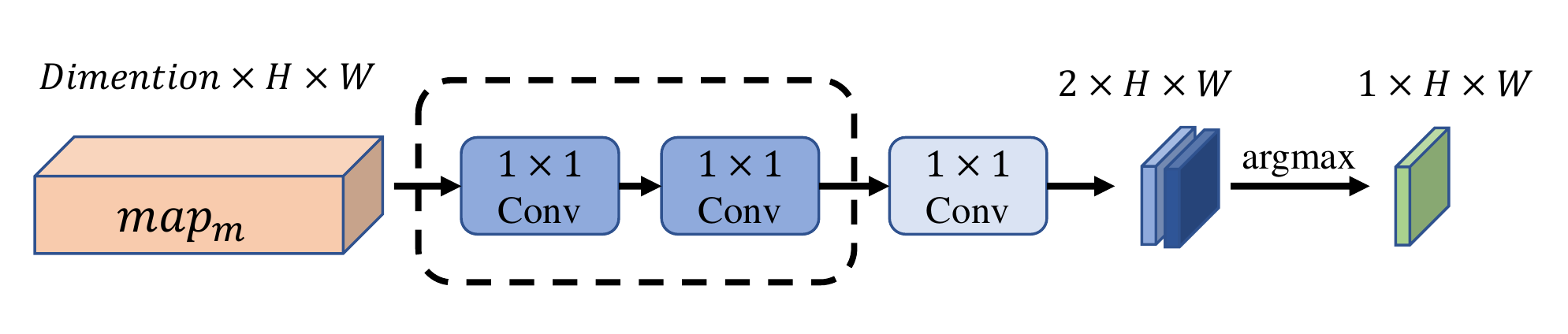}
\caption{The structure of AtF. We use two convolution layers in dotted box as shown on CUB-200, while use three convolution layers on ImageNet-1k. All the convolution layers have the same dimension as $map_m$.} \label{afhead}
\end{figure}

\section{Methodology}
In this section, firstly we introduce the main architecture of CaFT. Then, we analyze each module of CaFT respectively.

\subsection{Overview}
We use the ViT model as backbone and directly use the output of backbone as final classification result. 

We take the tokens of last three blocks out as $map_0$,$map_1$,$map_2$ respectively, and the position embedding parameter of ViT as $map_p$. Then we merge these four maps to $map_m$ as
\begin{equation}
   map_m=\sum_{i\in \left \{ 0,1,2,p \right \}} \alpha _i \times  map_i
   \label{mapmerge}
\end{equation}
where $\alpha _i$ is the merge ratio.

In training process, we have designed to have three steps on $map_m$, as is shown in Fig.\ref{structure}. 

\begin{enumerate}
    \item[(a)] At the beginning of training, we input the merged tokens of $map_m$ to the clustering module, in which these tokens are clustered into 3 categories by k-means algorithm and the tokens with the same category as class token are chosen to be foreground. The next step is to binarize the category distribution map according to the foreground and utilize classical image filter to reduce some noises. Up to this point, the initial masks of images have been generated.
    \item[(b)] The next step is to take the initial masks of Clustering as pseudo targets for Attention Filter 1 (AtF$_1$), which is a two-category classifier of foreground and background.
    \item[(c)] After finishing training of AtF$_1$, CaFT divides the input image into four parts, inputs them into the model respectively and merges their masks together to a more refined mask. A new AtF$_2$ with the same structure as AtF$_1$ is trained on the pseudo label of refined masks.
\end{enumerate}

In inference, the $map_m$ is directly input to AtF$_2$, and the output mask of AtF$_2$ is used to generate one bounding box for one image.
 
\subsection{ViT Backbone}
ViT firstly separates the image into patches, then these patches are flattened and put through a linear layer to become n tokens. Different from CNN network, ViT adds an extra class token and input it to network with other tokens from images. In addition, ViT adds a position embedding for every tokens before being input. Experiments of \cite{vit} show that position embedding is related to the corresponding position of tokens. The major structure of ViT is encoder, in which tokens will go through cascaded blocks consisted of a multi-head self-attention layer and a Multilayer Perceptron (MLP). After propagation in encoder, the network outputs the same number of tokens {t$_*$,t$_1$,…,t$_{n-1}$,t$_n$}, and uses the class token t$_*$ to inference the category.

A little different from original ViT model, we reserve output of the last three blocks of ViT for the subsequent process.

\subsection{Clustering of Tokens}
\begin{figure}
\begin{center}
    \includegraphics[width=0.8\textwidth]{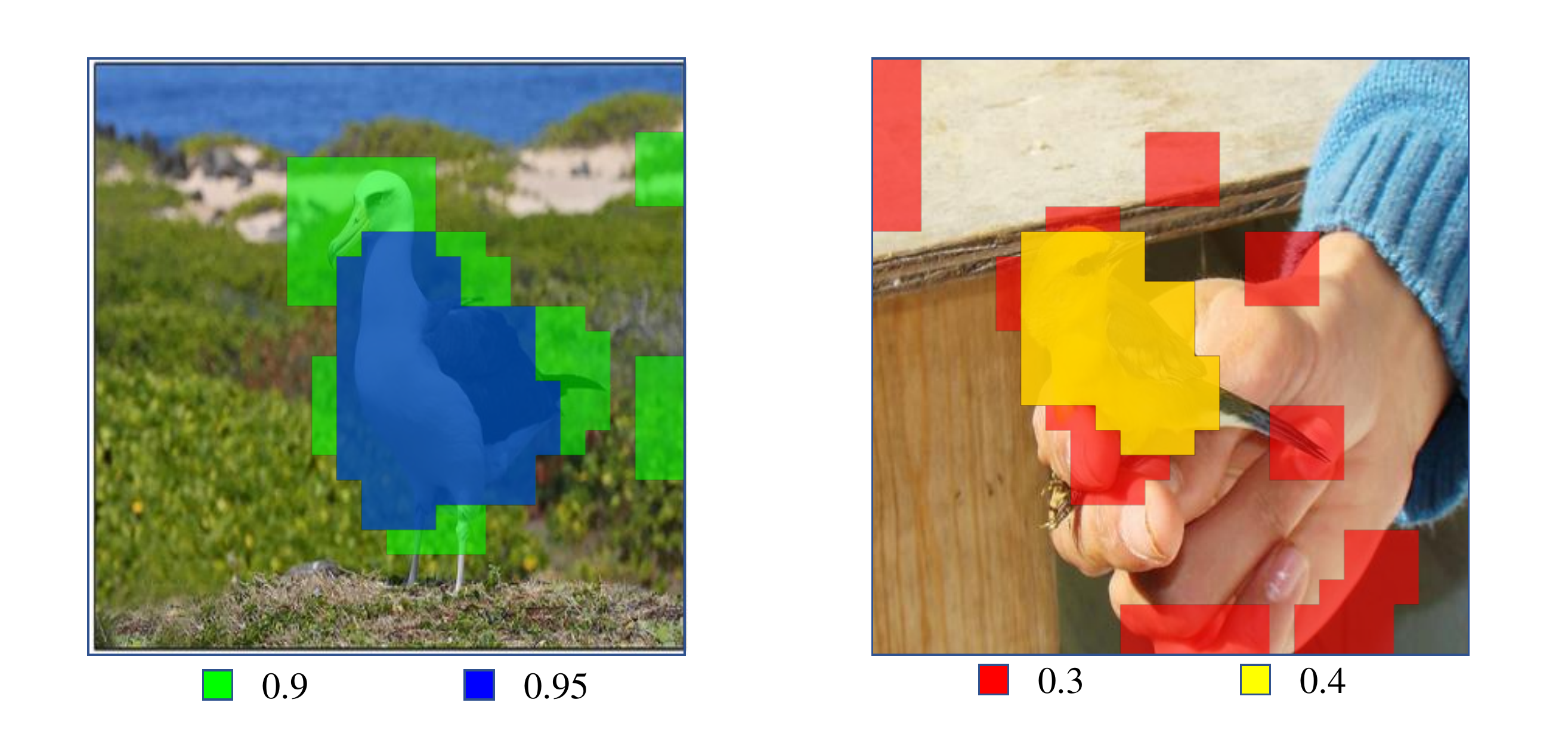}
\caption{Region of different threshold. In original CAM, after computing the classification scores of positions, it needs to choose a uniform threshold for different images. However, in the left image, the suitable threshold to cover object is around 0.9, while it is around 0.4 in the right image.} 
\label{thres}
\end{center}
\end{figure}

Classical CAM-based methods usually need to set a handcrafted fixed threshold to decide the foreground and background. For different inputs, the most suitable threshold to select foreground are different, as shown in Fig.\ref{thres}. However, under the previous frameworks, it is able to only choose a compromised value according to the performance on the whole dataset. In other words, a fixed design cannot extract the accurate boundary of foreground and background. Because of the compromised boundary, the predicted region is unable to cover the entire object, so it is inclined to focus on the most discriminative region.

Therefore, we decide to avoid this handcraft process. We find the eigenvectors that make up the feature map have apparent correlation with each other. But the similarity is still a continuous value and has different magnitude among images. 

\begin{figure}
\includegraphics[width=\textwidth]{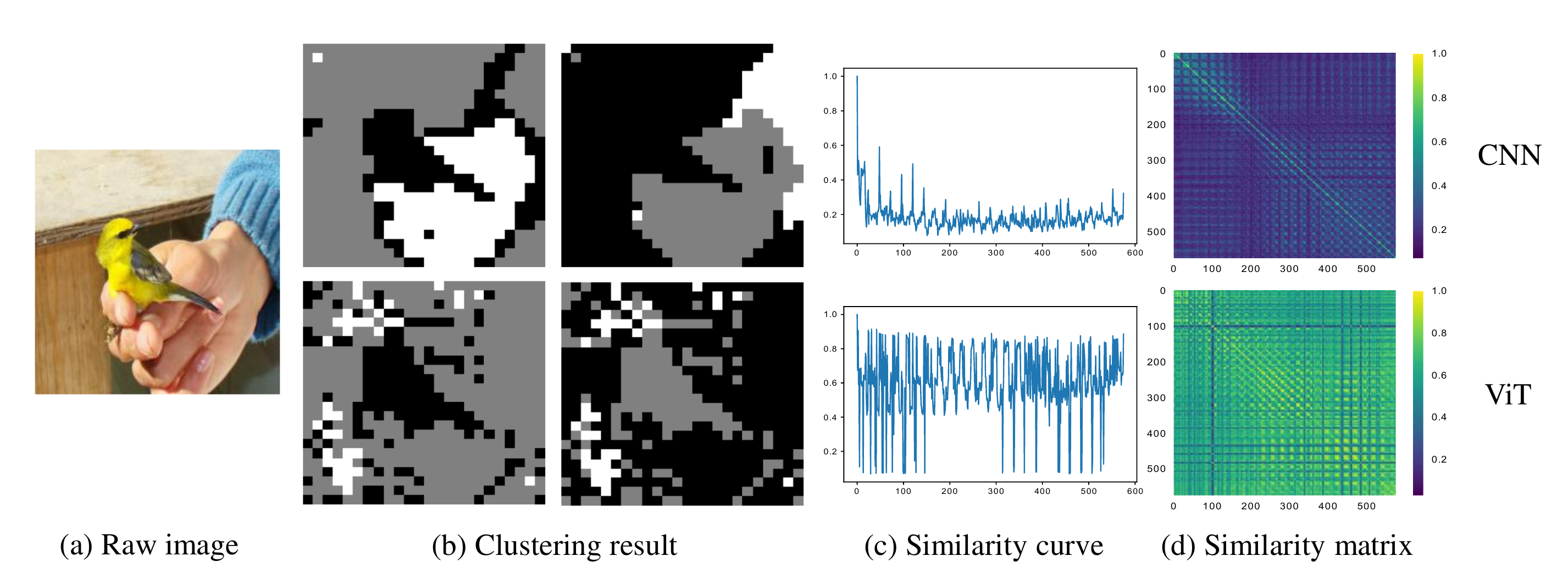}
\caption{Comparison between CNN and ViT. The top line is result of CNN and the second line is result of ViT. (a) The Raw image input. (b) The clustering result with different random seeds. (c) The similarity curve of points on the feature map. (d) The similarity matrix of points.} 
\label{cnnvitcompare}
\end{figure}

So, we apply clustering method to this task and choose the common k-means algorithm to automatically learn the boundary of foreground and background on the one-image level. At the beginning of experiment, as shown in Fig.\ref{cnnvitcompare}, we use the feature map of CNN-based network (we use ResNet50 \cite{resnet} in this experiment) to cluster category of each point. However, we find the result of clustering is unstable with different random seeds which influence the initial state of clustering, as shown in the top line in Fig.\ref{cnnvitcompare}(b). By computing the similarity matrix of eigenvectors, we find the similarity between vectors of CNN are commonly low, in Fig.\ref{cnnvitcompare}(c). Moreover, in Fig.\ref{cnnvitcompare}(d), we can see the vector tends to have higher similarity with the vectors among it and lower similarity with the vectors far from it even if they are all regions of object. Most important of all, there is a fatal problem of CNN that although we can cluster the categories, we have no way to decide which cluster belongs to the object’s region. 

\begin{figure}
\begin{center}
    \includegraphics[width=0.9\textwidth]{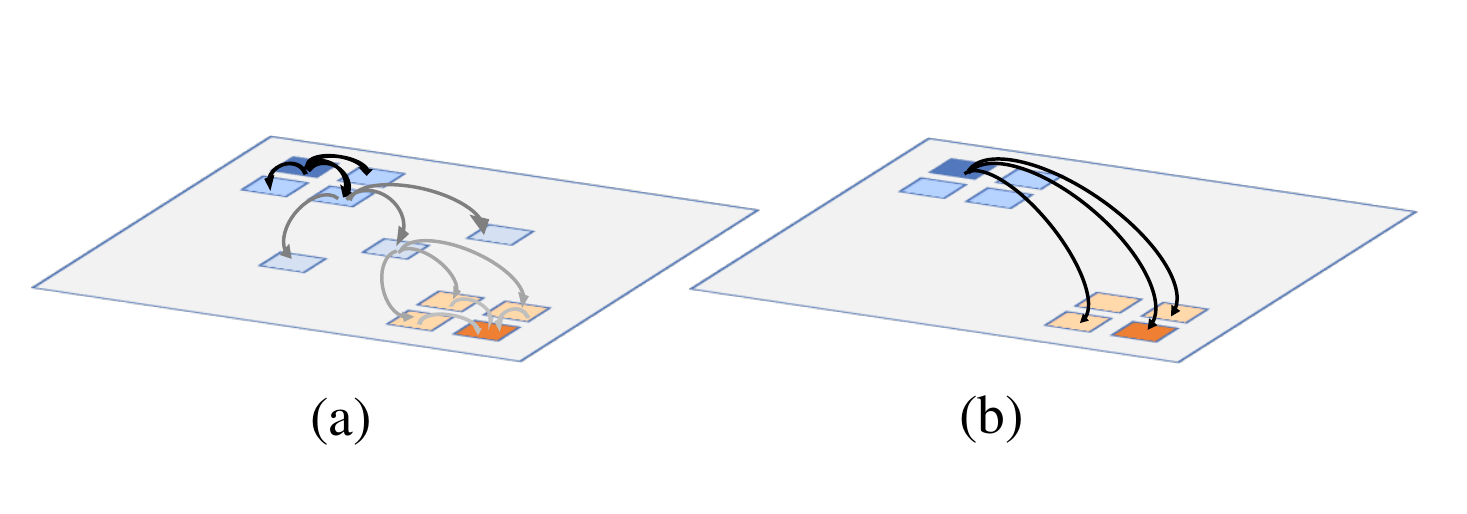}
\caption{The correlation of points in (a) CNN and (b) ViT.} \label{relation}
\end{center}
\end{figure}

It is easy to solve above problems by the Vision Transformer (ViT). Firstly, due to the special design of the extra class token, we find there are obvious correlations between class token and other tokens covering the object. 
And these correlations have a characteristic of long range, which is because of the different topological structure of CNN’s eigenvectors and ViT’s tokens. As is shown in Fig.\ref{relation}, eigenvectors are indirectly connected, while tokens acquire direct information from each other. From the similarity curve with the same image in Fig.\ref{cnnvitcompare}(c), it is obvious that the similarity of tokens has bigger variance than CNN’s vectors, which means a higher degree of difference between clustering classes. Therefore, the result of clustering is more stable for ViT's tokens.

To verify the feasibility of using class token as flag of object cluster, we compute the mean Euclidean distance between class token and center of clustering of sampling images. It (10.653) is close to the mean distance within cluster to center of clustering ($D_{ic}$) and is much less than clustering radius ($D_r$) of $map_m$ shown in Table \ref{clusterperform}. This proves that the class token belongs to the cluster of object region. 

As the Fig.\ref{result}(c) shows, after choosing the cluster of class token as positive region, the positive mask has almost covered the entire object but there are some noisy points in the background. So, we use simple smoothing filter of picture processing to suppress these noises, which outputs Fig.\ref{result}(d). Although the mask covers the entire object, the edge of mask does not perfectly fit to the object. Therefore, we proposed the next step of Attention Filter to fine tune the edge result and replace the clustering to accelerate the inference.

\subsection{Attention Filter}
Some CAM-based methods usually separate the training of WSOL into several steps. PSOL \cite{psol} uses the DDT \cite{ddt} algorithm to generate bounding boxes and considers these boxes as pseudo box-labels to train the box-regression network in the second stage. Inspired by this, we have designed a multi-stage training process. Different from regression of bounding boxes, we regard the subsequent training as segmentation training and take the output mask of clustering as pseudo label. 

Through the previous clustering we obtain a set of masks covering the object but have some noises of background. We use Gaussian filter on the noisy mask to smooth part of these points. After that, it already has the ability to predict the bounding box, and the model with backbone ViT-B can achieve 70.16\% and 57.59\% GT-Known accuracy on CUB-200 and ImageNet-1k respectively. 

However, it is not enough for a high-quality WSOL model. Besides the noise of background, the clustering process has a low inference speed because of the iteration of clustering algorithm. According to our early experiments on CNN, we find that there is less noise in the clustering result of CNN. Besides, convolution on the feature map with the same kernel is able to increase the resolving power of foreground and smooth the noise. Therefore, we use a shallow convolution head (Fig.\ref{afhead}) following the backbone, which directly extracts the attention mask of the object from merged tokens $map_m$. This module plays an analogous role as the filter in image denoising, so we named it as \textbf{At}tention \textbf{F}ilter (AtF). 

Take the ViT-B-384 for example, after obtaining $map_m$, we pull the class token out and send the rest as feature map with shape (D, H, W) to cascaded convolution layers (AtF$_1$) which outputs an attention mask with shape (2, H, W).  D means the embedding dimension of ViT. The 2 channels of output feature map correspond to two classes as background and foreground, and after choosing the points of foreground, AtF$_1$ outputs the (1, H, W) mask which could predict the bounding box of object by finding the external border. During training AtF$_1$, the ViT backbone is frozen, which can prevent the classification accuracy of backbone from being influenced by training AtF$_1$ and also avoid the ViT backbone with strong fitting ability from over-fitting the training set and the noise of the pseudo label.

\begin{figure}
\includegraphics[width=0.98\textwidth]{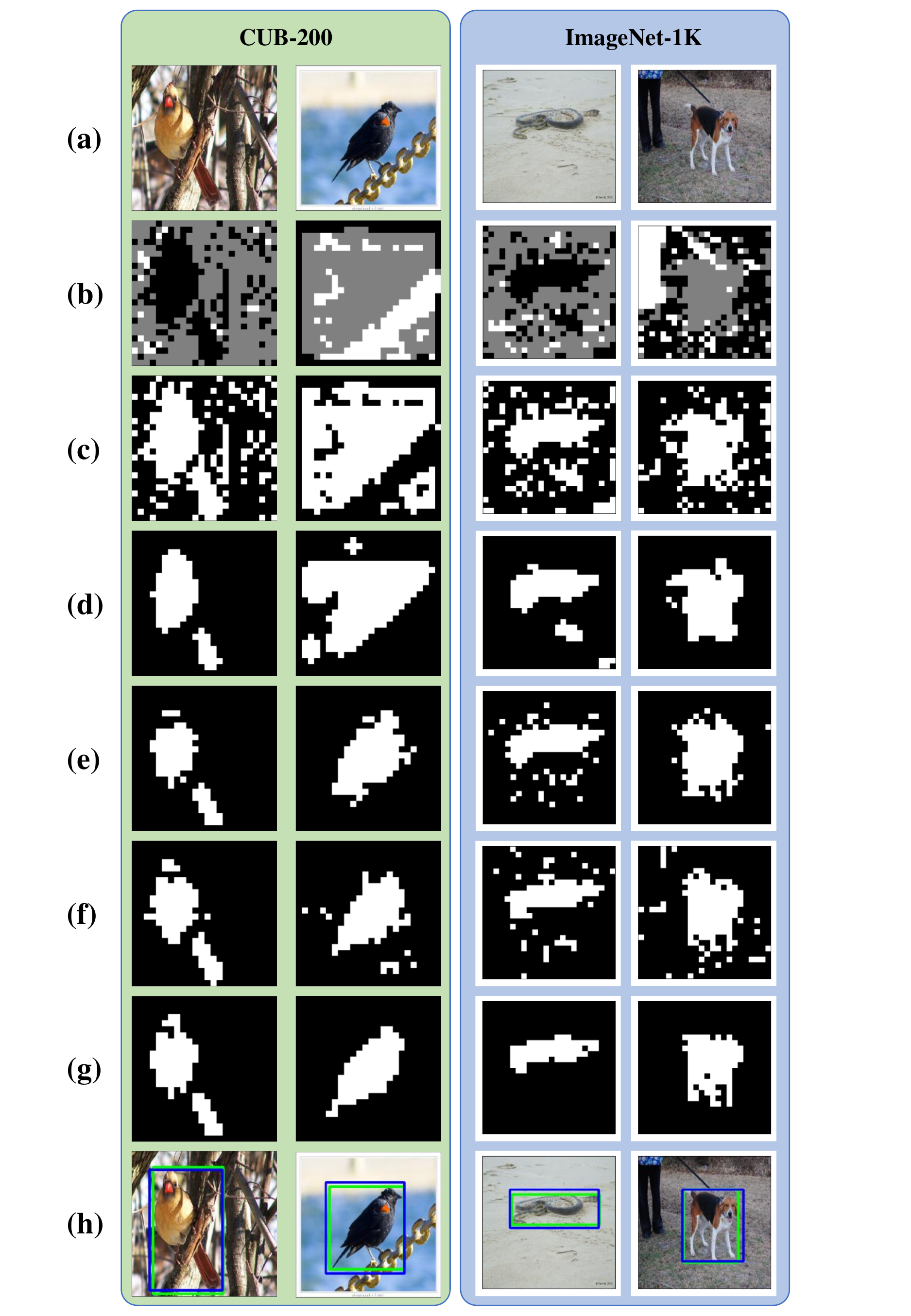}
\caption{Results of CaFT. (a) The raw images. (b) The clustering results. (c) Binarization according to class token. (d) The result of clustering. (e) The result of AtF$_1$. (f) The refined mask. (g) The result of AtF$_2$. (h) The ground truth box (green) and the prediction box (blue).} 
\label{result}
\end{figure}

After being trained on the initial mask from clustering, AtF$_1$ can achieve a big rise on location accuracy, from 70.16\% to 85.64\% on CUB-200. The output mask is shown in Fig.\ref{result}(e). The $1\times1$ convolution is similar to fully connected layer for each token, so it is essentially eigenspace transformation of tokens to enhance the feature. After transformation, it is able to separate the vague vectors around the cluster boundary. We do the same clustering process as original tokens of $map_m$ on feature maps of convolution layers in AtF (Conv1, Conv2). Table \ref{clusterperform} shows the comparison of following metrics. 
The mean within-cluster distance to center of clustering ($D_{ic}$), as 
\begin{equation}
    D_{ic}=\frac{1}{n_o} \sum_{x \in C_o} {\rm Euclidean} (x, c_o) ,
    \label{Dic}
\end{equation}
where Euclidean(*, *) is computing Euclidean distance, $C_o$ is the set of points in cluster of object region with size $n_o$, and $c_o$ is the center of cluster of object region.
Clustering radius ($D_r$), as
\begin{equation}
    D_r={\rm max}\left \{ {\rm Euclidean}(x, c_o) | x \in C_o \right \}.
    \label{Dr}
\end{equation}
Distance between centers of clustering ($D_{cc}$) and $D_{cc}$ divided by $D_{ic}$ and $D_r$, as
\begin{equation}
    D_{cc}=\frac{1}{k-1} \sum_{q=1}^{k} {\rm Euclidean}(c_o, c_q) ,
    \label{Dcc}
\end{equation}
where $k$ is the number of clusters in k-means algorithm, $c_q$ is the center of cluster $q$.
Calinski-Harabasz score \cite{score} ($Score$), which is defined as 
\begin{equation}
    Score=\frac{tr(B_k)}{tr(W_k)} \times \frac{n_E-k}{k-1},
    \label{Score}
\end{equation}
where $n_E$ is the size of the whole tokens set $E$, $W_k$ is the trace of the within-cluster dispersion matrix and $B_k$ is trace of the between group dispersion matrix defined by 
\begin{equation}
    W_k=\sum_{q=1}^{k}\sum_{x \in C_q}(x - c_q)(x-c_q)^T,
    B_k=\sum_{q=1}^{k}n_q(c_q-c_E)(c_q-c_E)^T
    \label{WkBk}
\end{equation}
with $c_E$ is the center of $E$ and $n_q$ is the number of points in cluster $q$. Calinski-Harabasz score evaluates the quality of clustering by within-cluster and among-cluster variance. A higher Calinski-Harabasz score relates to a model with better defined clusters.

\begin{table}
\caption{Comparison of clustering on $map_m$ and feature maps of AtF$_1$}
\begin{center}
\setlength{\tabcolsep}{2mm}{
\begin{tabular}{c|ccc|cc|c}
\hline \hline
Feature maps & $D_{ic}$    & $D_r$     & $D_{cc}$    & $D_{cc}/D_{ic}$ & $D_{cc}/D_r$ & $Score$    \\ \hline
$map_m$         & 9.29 & 18.03 & 16.58 & 1.78   & 0.92  & 188.86  \\
Conv1        & 8.37 & 16.16 & 24.72 & 2.95   & 1.53  & 688.55  \\
Conv2        & 3.87 & 9.30  & 19.39 & 5.01   & 2.08  & 1603.12 \\ \hline \hline
\end{tabular}}
\end{center}
\label{clusterperform}
\end{table}

\begin{figure}
\begin{center}
   \includegraphics[width=\textwidth]{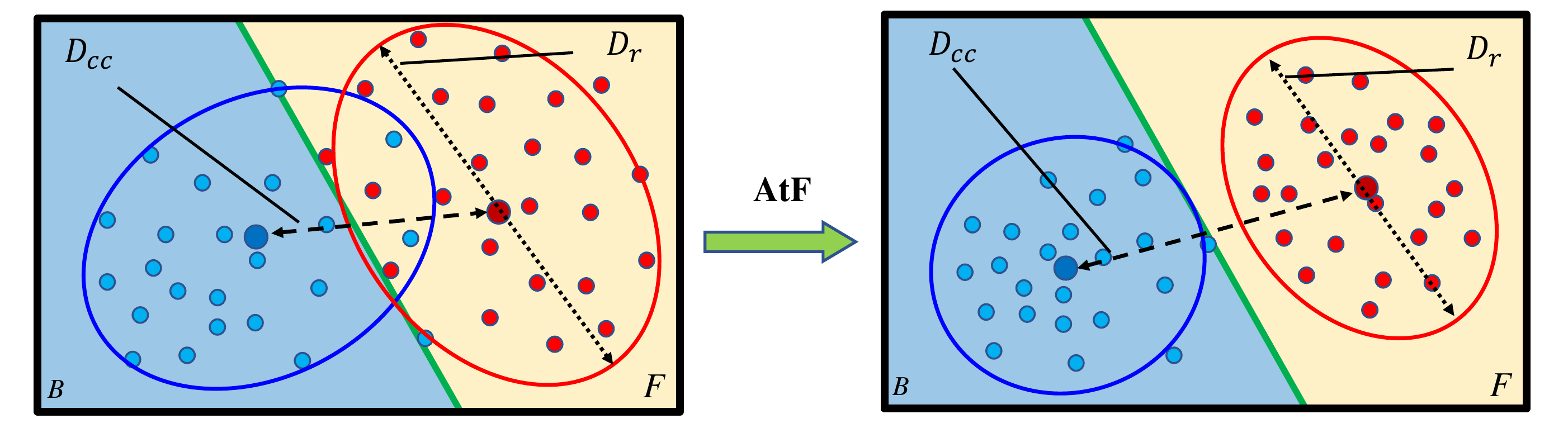}
\caption{Comparison of clustering results with and without AtF. } 
\label{clustershow} 
\end{center}
\end{figure}

Through AtF$_1$, $D_{ic}$ and $D_r$ descend, $D_{cc}/D_{ic}$, $D_{cc}/D_r$ and $Score$ increase, which means the degree of within-cluster aggregation is higher, the spacing between clusters is larger and a more stable clustering result. The visual result is shown in Fig.\ref{clustershow}. Therefore, AtF is able to enhance the feature and reduce the noisy points in the mask.

To further enhance the result, we divide the training image into four parts and use the AtF$_1$ to predict four masks. Then these four masks are stitched into a more refined mask which is the pseudo label for AtF$_2$. Compared to directly using the result mask of AtF$_1$, AtF$_2$ trained on the refined mask performs much better. Although the box prediction on refined mask is not much better on CUB-200 (even worse on ImageNet-1K), in Table \ref{refined mask}. Because the detail of edge preforms better in refined mask while there are some extra sporadic noisy points, which is shown in Fig.\ref{result}(f). However, as Fig.\ref{result}(g) shows, AtF$_2$ is able to smooth these noisy points and obtain a more accurate edge information from refined masks at the same time. While the AtF$_2$ trained on AtF$_1$ mask tends to drop into the local minimum as AtF$_1$ because they have similar structure. These are reasons why AtF$_2$ trained on refined mask has a big rise from 87.11\% to 94.17\% and surpasses the AtF$_2$ trained on AtF$_1$ mask by a large margin.

% \usepackage{multirow}
\begin{table}
\caption{Comparison between AtF$_1$ and refined mask on CUB-200}
\begin{center}
\setlength{\tabcolsep}{4mm}{
\begin{tabular}{l|cc|cc}
\hline \hline
\multicolumn{1}{c|}{\multirow{2}{*}{AtF$_2$ trained on}} & \multicolumn{2}{c|}{Own result} & \multicolumn{2}{c}{AtF$_2$ result} \\ \cline{2-5} 
\multicolumn{1}{c|}{}                        & Gt-Known  & mean IoU & Gt-Known       & mean IoU       
\\ \hline
AtF$_1$ mask                                         & 85.64         & 0.6629   & 86.97               & 0.6744         \\ 
Refined mask                                  & 87.11         & 0.6885   & 94.17               & 0.7418         \\ \hline \hline
\end{tabular}}
\end{center}
\label{refined mask}
\end{table}

By the way, We can only use the divide-method after obtaining AtF$_1$ rather than when clustering, because the divided part of images may not contain objects and the result of clustering from blank part is unauthentic.

% \begin{figure}
% \includegraphics[width=\textwidth]{result.pdf}
% \caption{Results of CaFT. (a) The raw images. (b) The clustering results. (c) Binarization according to class token. (d) The result of clustering. (e) The result of AtF. (f) The ground truth box (green) and the prediction box (blue).} \label{result}
% \end{figure}

\section{Experiments}
\subsection{Experimental Settings}
\subsubsection{Datasets.}
There are two accepted datasets to evaluate the performance of WSOL methods, CUB-200 \cite{cub} and ImageNet-1K \cite{imagenet}. CUB-200 contains 200 categories of birds with 5994 training images and 5794 testing images. ImageNet-1K is an outstanding classification dataset with 1000 classes, containing 1,281,197 training images and 50000 validation images. There are also the bounding boxes labels of objects on validation images.

\subsubsection{Evaluation Metrics.}
Following previous work \cite{cam}, Top-1/Top-5 localization accuracy (\emph{Top-1}/\emph{Top-5} \emph{Loc}) and localization accuracy with ground-truth class (\emph{Gt-Known Loc}) are applied as evaluation metrics. \emph{Gt-Known Loc} is correct when the intersection over union (IOU) between the predicted bounding box and the ground truth bounding box is 0.5 or more. \emph{Top-1 Loc} is correct when the Top-1 classification result and Gt-Known Loc are both correct. \emph{Top-5 Loc} is correct when given the Top-5 predictions of groundtruth labels and bounding boxes, there is one prediction which the classification result and localization result are both correct. Because our CaFT just predicts one box for one image, \emph{Top-5 Loc} is just different from \emph{Top-1 Loc} on the classification result.

\subsection{Implementation Details}
CaFT is based on the ViT backbone \cite{vit}, which is pre-trained on the ImageNet-1K \cite{imagenet}. Each input image is directly resized to (384, 384). For experiment on CUB-200, we use the pre-trained weight of ViT to train the classification backbone on CUB-200, the following training never changes parameters of the backbone. The top-1 classification accuracy of the backbone is around 88.50\% and 83.70\% on CUB-200 and ImageNet-1K respectively.

We use k-means algorithm in package scikit-learn, and set 3 categories as the target with other parameters in default. On CUB-200 we use the whole training set of 5994 images. For experiment on ImageNet-1k, because of the high computing pressure of k-means algorithm, we just randomly sample 10 images from each class in training set to generate pseudo mask label, and train the AtF on this mini set (10000 images in total). 

We use SGD optimizer for all training. For training AtFs, we set the learning rate as 0.1 with cosine scheduler and 20 epochs for CUB-200, and 5 epochs for ImageNet-1K. 

We have tested the results of ViT-B and ViT-L model. For ViT-B model, we follow the integral process with clustering, AtF$_1$, refined mask and AtF$_2$; while for ViT-L, we directly train the AtF$_2$ using the pseudo mask labels generated by ViT-B model. 

% \usepackage{multirow}
\begin{table}
\caption{Results at different stages of CaFT. Center cropping is not used.}
\begin{center}
\setlength{\tabcolsep}{1.5mm}{
\begin{tabular}{c|c|cc|cc}
\hline \hline
\multirow{2}{*}{Method} & \multirow{2}{*}{Backbone} & \multicolumn{2}{c|}{CUB-200}                  & \multicolumn{2}{c}{ImageNet-1K}               \\ \cline{3-6} 
                        &                           & \multicolumn{1}{c|}{Top-1 Loc} & Gt-Known Loc & \multicolumn{1}{c|}{Top-1 Loc} & Gt-Known Loc \\ \hline
Clustering              & ViT-B                     & \multicolumn{1}{c|}{63.63}     & 70.16        & \multicolumn{1}{c|}{49.68}     & 57.59        \\
AtF$_1$                    & ViT-B                     & \multicolumn{1}{c|}{77.32}     & 85.64        & \multicolumn{1}{c|}{54.74}     & 63.45        \\
Refined                 & ViT-B                     & \multicolumn{1}{c|}{78.62}     & 87.11        & \multicolumn{1}{c|}{54.04}         & 62.59            \\
AtF$_2$                    & ViT-B                     & \multicolumn{1}{c|}{84.79}     & 94.17        & \multicolumn{1}{c|}{56.02}     & 64.79        \\ \hline \hline
\end{tabular}}
\label{stages}
\end{center}
\end{table}

\subsection{Performance}
\subsubsection{Main Results.}
The main results of CaFT on CUB-200 and ImageNet-1K are shown in Table \ref{resutloncub} and Table \ref{resutlonimagenet}, compared with other methods. The large model achieves the best result on metrics and the base model also has a competitive performance. There is an interesting phenomenon that \emph{Gt-Known Loc} divided by \emph{Top-1 Loc} surpasses 90.00\% on CUB-200 and 86.50\% on ImageNet-1K, which surpasses the Top-1 classification accuracy of the backbone (around 88.50\% on CUB-200 and 83.70\% on ImageNet-1K). A reasonable explanation is that the location accuracy and classification accuracy of CaFT have a significant positive correlation, while there is often a contradiction between this two in CAM. The attention of CaFT will not be payed more to discriminative region along with the increase of classification accuracy; on the contrary, the mask generated by a better classification backbone will be consisted of fewer noisy points, which benefits the clustering and training of AtF, as well as localization. 
The \emph{Top-1 Loc} and \emph{Gt-Known Loc} of different stages of CaFT are shown in Table\ref{stages}.

% \usepackage{multirow}
\begin{table}
\caption{Comparison of CaFT with state-of-the-art methods on the CUB-200}
\begin{center}
\begin{threeparttable}
\begin{tabular}{c|c|ccc}
\hline \hline
\multirow{2}{*}{Methods} & \multirow{2}{*}{Backbone}   & \multicolumn{3}{c}{CUB-200}                                                                \\ \cline{3-5} 
                         &                             & \multicolumn{1}{c|}{Top-1 Loc}       & \multicolumn{1}{c|}{Top-5 Loc}       & Gt-Known Loc   \\ \hline
CAM \cite{cam}                      & VGG-GAP                     & \multicolumn{1}{c|}{36.13}          & \multicolumn{1}{c|}{-}              & -              \\
ACoL \cite{acol}                     & VGG-GAP                     & \multicolumn{1}{c|}{45.92}          & \multicolumn{1}{c|}{56.51}          & 62.96          \\
ADL \cite{adl}                      & VGG-GAP                     & \multicolumn{1}{c|}{52.36}          & \multicolumn{1}{c|}{-}              & 73.96          \\
DDT \cite{ddt}                      & VGG16                       & \multicolumn{1}{c|}{62.3}           & \multicolumn{1}{c|}{78.15}          & 84.55          \\
SPG \cite{spg}                      & InceptionV3                 & \multicolumn{1}{c|}{46.64}          & \multicolumn{1}{c|}{57.72}          & -              \\
ADL \cite{adl}                      & ResNet50-SE                 & \multicolumn{1}{c|}{62.29}          & \multicolumn{1}{c|}{-}              & 71.99          \\
I2C \cite{i2c}                      & InceptionV3                 & \multicolumn{1}{c|}{65.99}          & \multicolumn{1}{c|}{68.34}          & 72.6           \\
DANet \cite{danet}                    & VGG16                       & \multicolumn{1}{c|}{52.5}           & \multicolumn{1}{c|}{62}             & 67.7           \\ \hline \hline
PSOL \cite{psol}                     & DenseNet161+EfficientNet-B7 & \multicolumn{1}{c|}{77.44}          & \multicolumn{1}{c|}{89.51}          & 93.01          \\
SPOL \cite{spol}                     & ResNet50+DenseNet161        & \multicolumn{1}{c|}{79.74}          & \multicolumn{1}{c|}{93.69}          & 96.46          \\
SPOL \cite{spol}                     & ResNet50+EfficientNet-B7    & \multicolumn{1}{c|}{80.12}          & \multicolumn{1}{c|}{93.44}          & 96.46          \\ 
TS-CAM \cite{tscam}                   & Deit-S                      & \multicolumn{1}{c|}{71.3}           & \multicolumn{1}{c|}{83.8}           & 87.7           \\
TS-CAM \cite{tscam}                   & Deit-B-384                  & \multicolumn{1}{c|}{75.8}           & \multicolumn{1}{c|}{84.1}           & 86.6           \\
TS-CAM \cite{tscam}                   & Conformer-S                 & \multicolumn{1}{c|}{77.2}           & \multicolumn{1}{c|}{90.9}           & 94.1           \\ \hline \hline
CaFT(ours)               & ViT-B-384\tnote{*}                  & \multicolumn{1}{c|}{84.79}          & \multicolumn{1}{c|}{92.75}          & 94.17          \\
CaFT(ours)               & ViT-B-384                   & \multicolumn{1}{c|}{86.57}          & \multicolumn{1}{c|}{94.1}           & 95.5           \\
CaFT(ours)               & ViT-L-384\tnote{*}                   & \multicolumn{1}{c|}{86.8}           & \multicolumn{1}{c|}{95.18}          & 96.44          \\
CaFT(ours)               & ViT-L-384                   & \multicolumn{1}{c|}{\textbf{88.26}} & \multicolumn{1}{c|}{\textbf{96.38}} & \textbf{97.55} \\ \hline \hline
\end{tabular}
\label{resutloncub}
\begin{tablenotes}
        \footnotesize
        \item[*] This model does not use center cropping.
      \end{tablenotes}
  \end{threeparttable}
\end{center}

\end{table}

% \usepackage{multirow}
\begin{table}
\caption{Comparison of CaFT with state-of-the-art methods on the ImageNet-1k}
\begin{center}
\begin{threeparttable}
\begin{tabular}{c|c|ccc}
\hline
\hline
\multirow{2}{*}{Methods} & \multirow{2}{*}{Backbone}   & \multicolumn{3}{c}{ImageNet-1k}                                                            \\ \cline{3-5} 
                         &                             & \multicolumn{1}{c|}{Top-1 Loc}       & \multicolumn{1}{c|}{Top-5 Loc}       & Gt-Known Loc   \\ \hline
CAM \cite{cam}                      & VGG-GAP                     & \multicolumn{1}{c|}{42.8}           & \multicolumn{1}{c|}{54.86}          & 59             \\
ACoL \cite{acol}                     & VGG-GAP                     & \multicolumn{1}{c|}{45.83}          & \multicolumn{1}{c|}{59.43}          & 62.96          \\
ADL \cite{adl}                      & VGG-GAP                     & \multicolumn{1}{c|}{44.92}          & \multicolumn{1}{c|}{-}              & -              \\
DDT \cite{ddt}                      & VGG16                       & \multicolumn{1}{c|}{47.31}          & \multicolumn{1}{c|}{58.23}          & 61.41          \\
SPG \cite{spg}                      & InceptionV3                 & \multicolumn{1}{c|}{48.6}           & \multicolumn{1}{c|}{60}             & 64.69          \\
ADL \cite{adl}                      & ResNet50-SE                 & \multicolumn{1}{c|}{48.53}          & \multicolumn{1}{c|}{-}              & -              \\
I2C \cite{i2c}                      & InceptionV3                 & \multicolumn{1}{c|}{53.11}          & \multicolumn{1}{c|}{64.13}          & 68.5           \\
DANet \cite{danet}                    & GoogLeNet                   & \multicolumn{1}{c|}{47.5}           & \multicolumn{1}{c|}{58.3}           & -              \\ \hline \hline
PSOL \cite{psol}                     & DenseNet161+EfficientNet-B7 & \multicolumn{1}{c|}{58}             & \multicolumn{1}{c|}{65.02}          & 66.28          \\
SPOL \cite{spol}                     & ResNet50+DenseNet161        & \multicolumn{1}{c|}{56.4}           & \multicolumn{1}{c|}{66.48}          & 69.02          \\
SPOL \cite{spol}                     & ResNet50+EfficientNet-B7    & \multicolumn{1}{c|}{59.14}          & \multicolumn{1}{c|}{67.15}          & 69.02          \\ 
TS-CAM \cite{tscam}                   & Deit-S                      & \multicolumn{1}{c|}{53.4}           & \multicolumn{1}{c|}{64.3}           & 67.6           \\ \hline \hline
CaFT(ours)               & ViT-B-384\tnote{*}           & \multicolumn{1}{c|}{56.02}          & \multicolumn{1}{c|}{63.49}          & 64.79          \\
CaFT(ours)               & ViT-B-384                   & \multicolumn{1}{c|}{58.06}          & \multicolumn{1}{c|}{65.74}          & 67.03          \\
CaFT(ours)               & ViT-L-384\tnote{*}        & \multicolumn{1}{c|}{58.35}          & \multicolumn{1}{c|}{65.91}          & 67.2           \\
CaFT(ours)               & ViT-L-384                   & \multicolumn{1}{c|}{\textbf{60.59}} & \multicolumn{1}{c|}{\textbf{68.56}} & \textbf{69.86} \\ \hline \hline
\end{tabular}
\label{resutlonimagenet}
\begin{tablenotes}
        \footnotesize
        \item[*] This model does not use center cropping.
      \end{tablenotes}
  \end{threeparttable}
\end{center}

\end{table}

\subsubsection{About Center Cropping.}
A lot of recent previous work \cite{acol,spg,psol,spol} used center cropping during inference, we follow these works and get the results of our CaFT to compare the result with them. However, we think center cropping is a method to resize the ground truth boxes and reduce the difficulty of the task regardless of model design. Because by cropping the input images, the proportion of the bounding box in the image is increased so that even if using the boundary of the image directly, the IoU will be higher than 0.5 and the prediction will be regarded as a positive sample. Obviously, this method makes the result of evaluation unreliable, so we also record the result without center cropping to show the objective results. 

Our CaFT outperforms the previous work on \emph{Top-1 Loc}, \emph{Top-5 Loc} and \emph{Gt-Known Loc} metrics. With ViT-L backbone, CaFT achieves 88.26\% \emph{Top-1 Loc} and 97.55\% \emph{Gt-Known Loc} on CUB-200, 60.59\% \emph{Top-1 Loc} and 69.86\% \emph{Gt-Known Loc} on ImageNet-1K. With ViT-B backbone, CaFT also achieves 86.57\% \emph{Top-1 Loc} and 95.50\% \emph{Gt-Known Loc} on CUB-200, 58.06\% \emph{Top-1 Loc} and 67.03\% \emph{Gt-Known Loc} on ImageNet-1K, which is also a competitive result. As the opinion in \cite{tscam}, transformer-based networks preserve the entirety view of objects. By clustering, CaFT extracts a more complete mask from feature map than CAM. Although there are noisy points, through the subsequent convolution layers with $1\times1$ kernel size, these points are easy to erase. 

\begin{figure}
\begin{center}
    \includegraphics[width=0.8\textwidth]{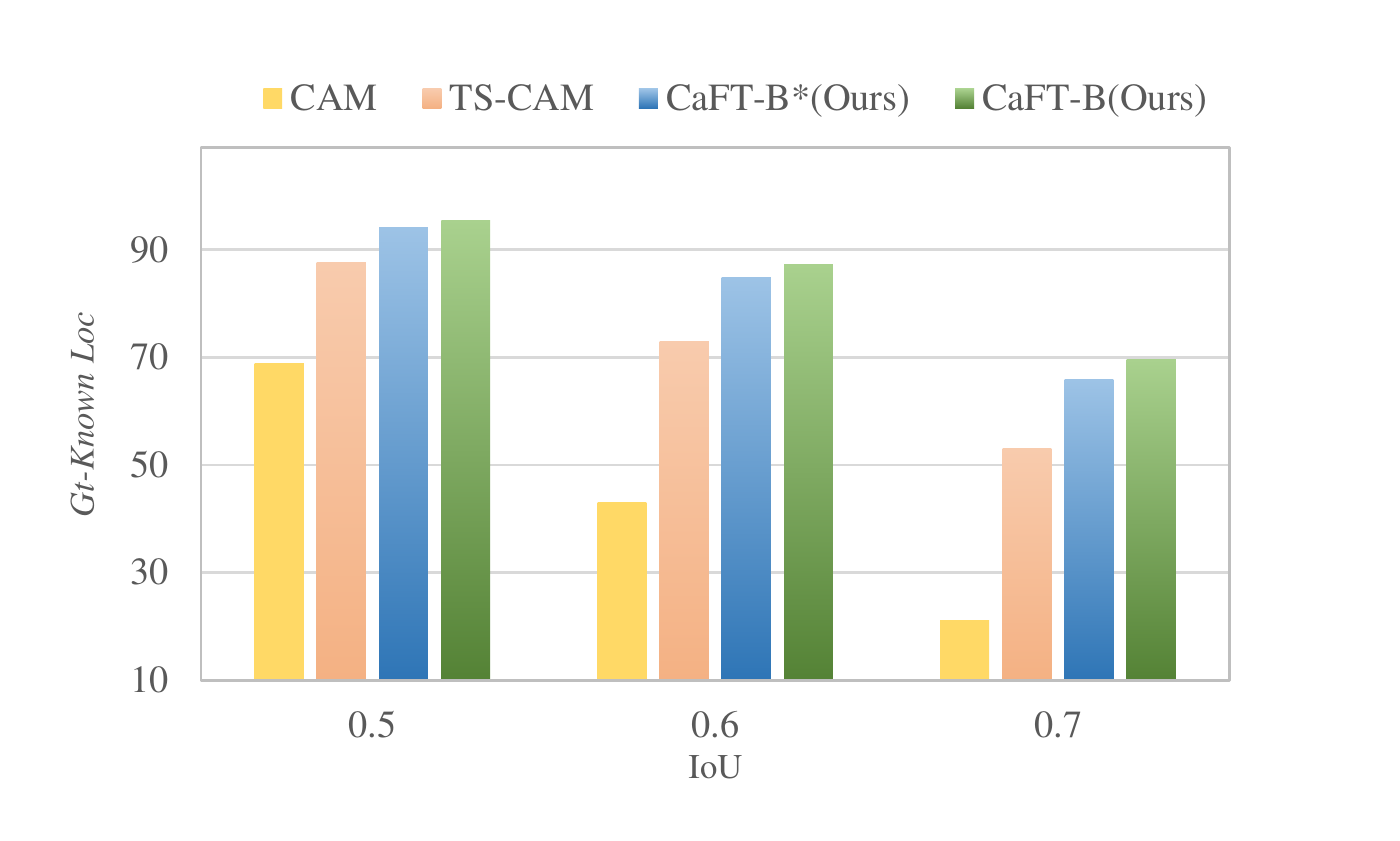}
\caption{Comparison of Gt-Known localization accuracy under different IoUs on CUB-200\cite{cub}. CaFT-B* does not use center cropping.} 
\label{multiiou}
\end{center}

\end{figure}

To compare the effectiveness of the model comprehensively, we show the localization under multiple IoUs in Fig.\ref{multiiou}. CaFT outperforms the previous methods in each IoU threshold.

\subsection{Ablation Study and More Details}
In this section, we will verify the components of CaFT by ablation study on CUB-200 dataset, and introduce the effect of some details. 

\subsubsection{Attention Filter.}
\begin{figure}
\includegraphics[width=\textwidth]{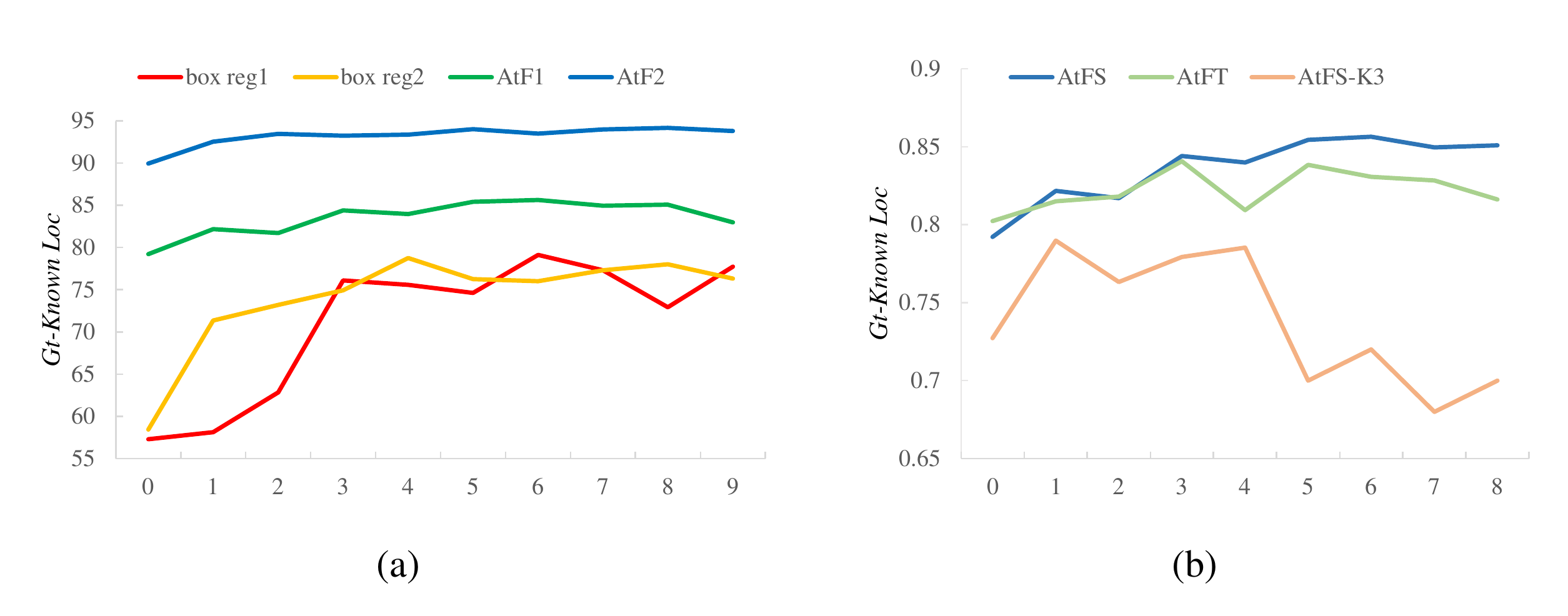}
\caption{Training process. (a) Comparison of the box regression and AtFs. (b) Comparison of different structure of AtF.} 
\label{trainprocess}
\end{figure}

Different from \cite{psol}, we use the masks from clustering to be pseudo labels rather than bounding boxes. To compare the effect of these two kinds of pseudo labels, we set the other structures and parameters fixed and replace the AtF in different stages by bounding boxes regression. The result and the process of training is recorded in Fig.\ref{trainprocess} (a) and Table\ref{trainingmethods}. 

The box reg1 is trained on the pseudo box labels generated by clustering and the box reg2 is trained on the pseudo box labels generated by AtF$_1$. The backbone of the box regression is ViT-B to avoid the effect of different backbones, and we replace the MLP head of class token by a regression head of [x, y, w, h], and the regression parameters are relative value of the image size. From the Table \ref{atf} we can see that even if being trained on the same result of clustering, box reg1 is unable to achieve the same location accuracy as AtF$_1$. Moreover, for box reg2 trained on the prediction boxes of AtF$_1$, the results fall down sharply. 
Besides the finally accuracy, the curves of the training process show that the convergence of box regression is worse than AtF. This experiment proves that the box regression is capable to increase the location accuracy to a certain extent, but is unable to obtain a stable and high-quality result. On the one hand, box regression usually needs a larger model than AtF and it is easy to over-fit the training set as well as the noise of pseudo labels. On the other hand, according to the experience in fully supervised location, box regression is an inherently unstable training method because of the four-parameter regression and it is more severe with pseudo labels. Therefore, box regression is powerless to smooth and correct the error in pseudo labels.

We try to directly use deep network to extract the mask from raw images with the pseudo mask labels from clustering. We use ResNet50 \cite{resnet} and VGG16 \cite{vgg} and the results are shown in Table\ref{trainingmethods}. However, it is really easy to over-fitting the noise of background for these networks due to the depth, so the location accuracy is even worse than it of Clustering.

\begin{table}
\caption{Comparison of different training methods on pseudo mask.}
\begin{center}
\setlength{\tabcolsep}{5mm}{
\begin{tabular}{c|c|c}
\hline \hline
Method   & Backbone & Gt-Known Loc \\ \hline
Clustering  & ViT-B & 70.16         \\ 
AtF$_1$     & ViT-B    & 85.64        \\
AtF$_2$     & ViT-B    & 94.17        \\
box reg1 & ViT-B    & 79.13        \\
box reg2 & ViT-B    & 78.75        \\
Direct   & ResNet50 & 68.42        \\
Direct   & VGG16    & 58.21        \\ \hline \hline
\end{tabular}}
\label{atf}
\end{center}
\label{trainingmethods}
\end{table}

A vital problem of pseudo labels is over-fitting, so we control the size of AtF and compare the effect of different structures in Fig.\ref{trainprocess}(b). AtFS has two convolution layers of $1\times1$ kernel with batch normalization and activation function; AtFT has one convolution layer of $1\times1$ kernel with batch normalization and activation function; AtFS-K3 has the same structure with AtFS with $3\times3$ kernel in the first convolution layer. The result shows that AtFS is the best structure with stable curve and there is over-fitting in AtFS-K3. The normalized result of trained parameters of $3\times3$ kernel shows that the center of the kernel has much more weight than the surrounding's. Therefore, the center is crucial while merging surrounding information has opposite effect which may confuse the judgement of the edge points.

\subsubsection{Fusion of Multi-layers.}
\begin{figure}
\includegraphics[width=\textwidth]{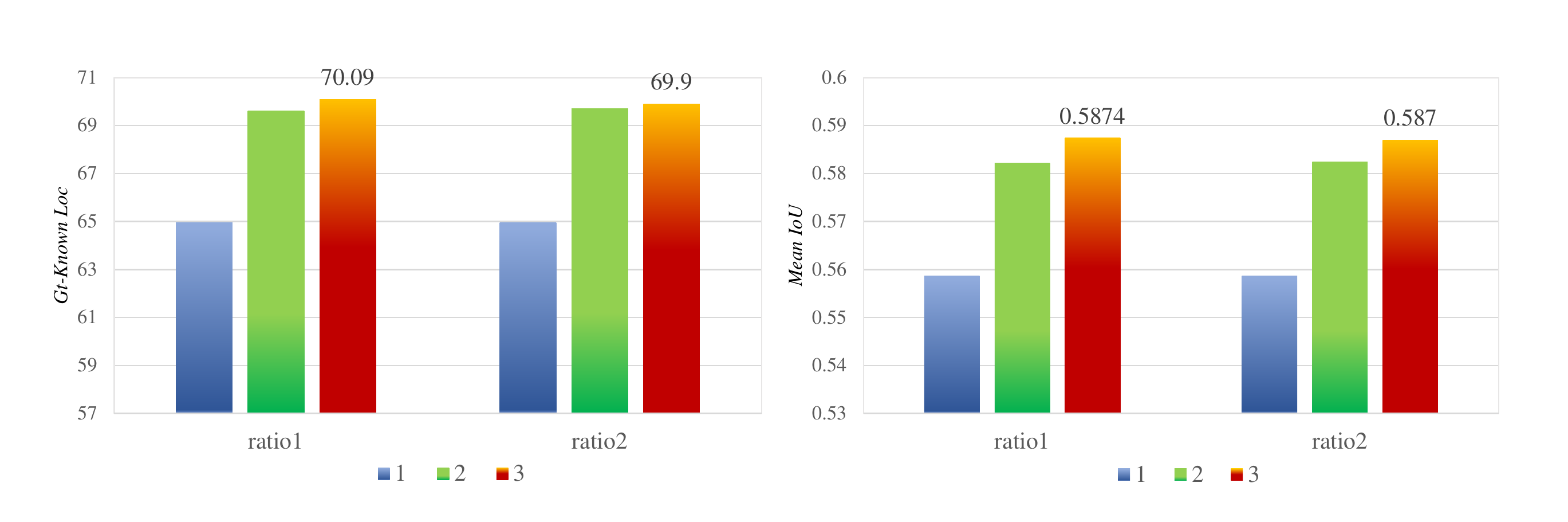}
\caption{Comparison of different number of layers to be merged. Respectively, "1, 2, 3" means the number of layers being merged.} 
\label{ablation}
\end{figure}

CaFT merges the last three layers of tokens and position embedding parameters before clustering and AtF. We compare the influence of the quantity of fusion layers to clustering result. As shown in Fig.\ref{ablation}, with different merge ratio, the more layers merged, \emph{Gt-Known Loc} and \emph{Mean IoU} both rise, but the magnitude of increase also declines. Moreover, a measure of emphasizing on the last layer results in a bit of increase. 

Merging position embedding parameters of ViT can also lift the result a little. Keeping the ratio fixed, results with position embedding parameters is higher 0.05\% than their counterparts. 

\section{Conclusion}
In this paper, we have proposed the Clustering and Filter on Tokens (CaFT) model for weakly supervised object location. CaFT aims to solve the problem of discriminative region preference and threshold choosing in WSOL. We use the clustering method and the special class token in Vision Transformer (ViT) backbone to generate the initial mask of object. To filter the noise of the initial mask, we train a shallow convolution head (Attention Filter, AtF) to extract a more accurate mask from tokens with the pseudo label of initial mask. Experiments on the CUB-200 and ImageNet-1K datasets show the effectiveness of CaFT. Different from CAM-based methods, CaFT provides a fresh way to think about WSOL tasks with clustering.

\bibliographystyle{splncs04}
\bibliography{Mybib}
\end{document}